\documentclass[aoas,accepted]{imsart}

\RequirePackage[OT1]{fontenc}
\RequirePackage{amsthm,amsmath}
\RequirePackage{natbib}
\RequirePackage[colorlinks,citecolor=blue,urlcolor=blue]{hyperref}

\arxiv{arXiv:1801.02858}

\startlocaldefs
\numberwithin{equation}{section}
\theoremstyle{plain}

\endlocaldefs

\usepackage{pdflscape}

\usepackage{bm}

\usepackage{color}
\usepackage{graphicx,psfrag,epsf}
\usepackage{hyperref}
\usepackage{enumerate}
\usepackage{natbib}
\usepackage{subcaption}
\let\cite\citep
\usepackage{colortbl}
\usepackage{soul}
\usepackage{url}
\usepackage{xcolor}
\usepackage{afterpage}

\usepackage{booktabs}
\usepackage{longtable}
\usepackage{makecell}

\usepackage{cellspace}
\hypersetup{
    colorlinks,
    linkcolor={red!50!black},
    citecolor={blue!50!black},
    urlcolor={blue!80!black}
}

\usepackage{bm}

\newcommand{\bs}{\bm{s}}
\newcommand{\bt}{\bm{t}}

\begin{document}
\begin{frontmatter}

\title{Scalable high-resolution forecasting of sparse spatiotemporal events with
kernel methods: a winning solution to the NIJ ``Real-Time Crime Forecasting Challenge''}

	\runtitle{Forecasting spatiotemporal events with kernel methods}
\begin{aug}
\author{\fnms{Seth} \snm{Flaxman}\thanksref{t1,m1}\ead[label=e1]{s.flaxman@imperial.ac.uk}},
\author{\fnms{Michael} \snm{Chirico}\thanksref{m2}\ead[label=e2]{}}, \\
\author{\fnms{Pau} \snm{Pereira}\thanksref{m3}\ead[label=e3]{}}
\and
\author{\fnms{Charles} \snm{Loeffler}\thanksref{m4}\ead[label=e4]{}}

\thankstext{t1}{Support was provided by the EPSRC (EP/K009362/1) and the European Research Council under the European Union's Seventh Framework Programme (FP7/2007-2013) ERC grant agreement no.~617071. Source code to reproduce our results is available:
\url{https://github.com/MichaelChirico/portland}.}
\affiliation{ Department of Mathematics and Data Science Institute, \\ Imperial College London\thanksmark{m1}, Grab, Singapore\thanksmark{m2}, Amazon, Inc.\thanksmark{m3} and Department of Criminology, University of Pennsylvania\thanksmark{m4} }

\address{Address of the corresponding author\\
\printead{e1}\\
\phantom{E-mail:\ s.flaxman@imperial.ac.uk}\printead*{e2}}

\end{aug}

\begin{abstract} We propose a generic spatiotemporal event forecasting method, which we developed for the National Institute of Justice's (NIJ) Real-Time Crime Forecasting Challenge~\cite{nij_real-time_2017}. Our method is a spatiotemporal forecasting model combining scalable randomized Reproducing Kernel Hilbert Space (RKHS) methods for approximating Gaussian processes with autoregressive smoothing kernels in a regularized supervised learning framework.  While the smoothing kernels capture the two main approaches in current use in the field of crime forecasting, kernel density estimation (KDE) and self-exciting point process (SEPP) models, the RKHS component of the model can be understood as an approximation to the popular log-Gaussian Cox Process model. For inference, we discretize the spatiotemporal point pattern and learn a log-intensity function using the Poisson likelihood and highly efficient gradient-based optimization methods. Model hyperparameters including quality of RKHS approximation, spatial and temporal kernel lengthscales, number of autoregressive lags, bandwidths for smoothing kernels, as well as cell shape, size, and rotation, were learned using crossvalidation.  Resulting predictions significantly exceeded baseline KDE estimates and SEPP models for sparse events.

\end{abstract}

\begin{keyword}
	\kwd{spatial statistics}
	\kwd{time series}
	\kwd{supervised learning}
	\kwd{spatiotemporal forecasting}
	\kwd{Cox process}
	\kwd{RKHS}
\end{keyword}

\end{frontmatter}

\section{Introduction}
\label{section:intro}

\begin{figure}[hb!]
	\begin{center}
\includegraphics[width=0.4\textwidth]{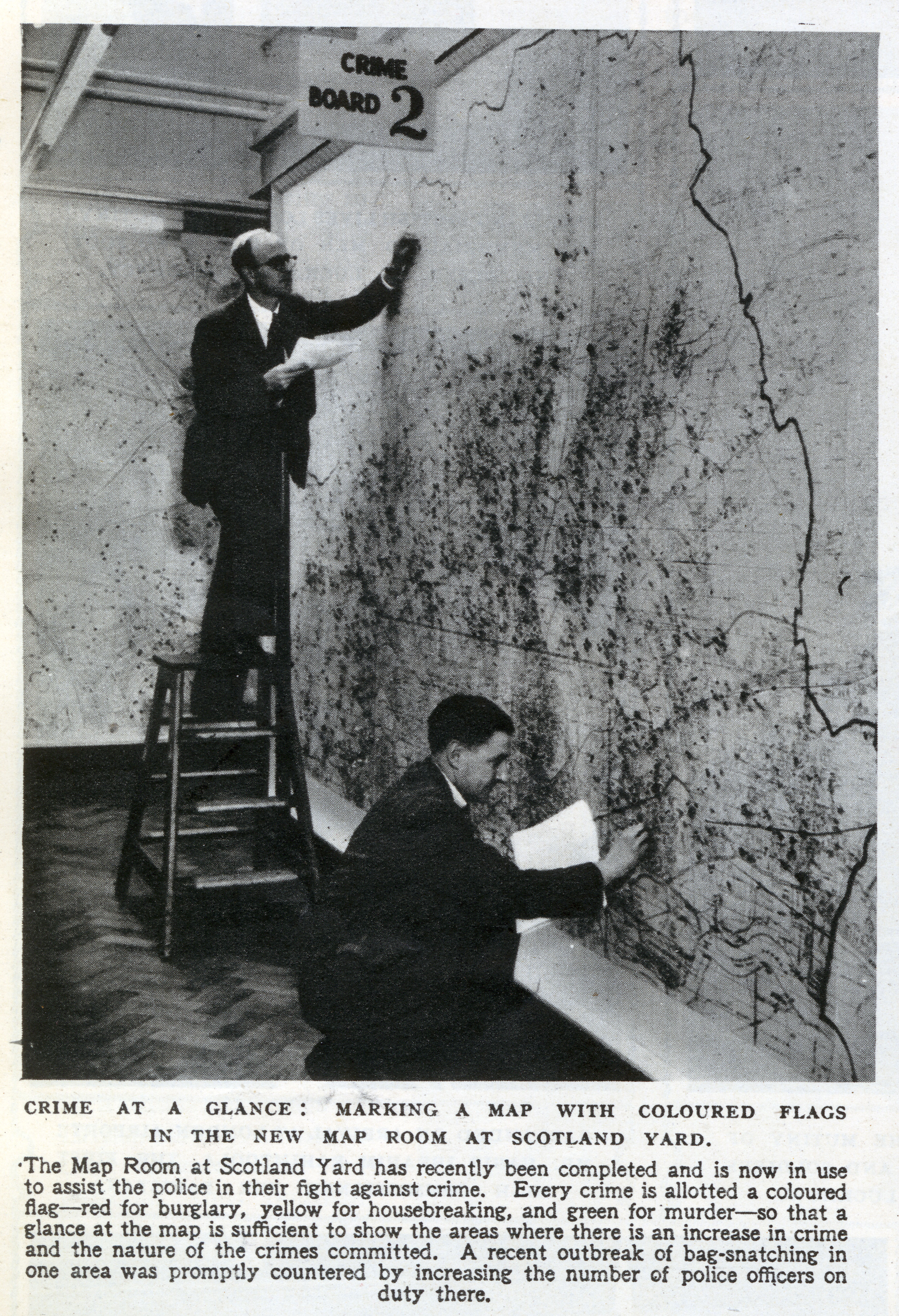}
\caption{Early use of crime pin-maps at Scotland Yard. 1947 \copyright Illustrated London News Ltd/Mary Evans}
\label{fig:Scotland_Yard}
	\end{center}
\end{figure}
Spatiotemporal forecasting of crime has been the focus of considerable
attention in recent years as academic researchers, police
departments, and commercial entities have all sought to build forecasting tools to predict when and where crimes are likely to occur \cite{perry_predictive_2013}. The earliest crime
forecasting tools consisted of nothing more than pin-maps (See 
Figure \ref{fig:Scotland_Yard}).  Prior week's crimes were mapped and qualitative
assessments of density, location, stability and significance were made
\cite{NCR4110110308}.

Subsequent tools have adopted a range of different smoothing techniques to
augment this method with kernel density estimation the most commonly used approach
\cite{gorr_early_2015,porter_evaluating_2012,chainey_utility_2008,johnson_predictive_2009}. Many methods are
model-driven, based on theories of crime causation
\cite{Caplan_2011,Mohler_2011}. Some use log-Gaussian Cox Processes 
(LGCPs) \cite{rodrigues2012bayesian,shirota2017space}, while others use
self-exciting point process models (SEPPs)
\cite{levine_crimestat:_2004,liu_criminal_2003,taddy_autoregressive_2010,Mohler_2011,rosser2016improving}
based on evidence of elevated levels of near-repeat victimization
\cite{pease1998repeat}. Some use additional
information, such as weather, demographics, and even social media
\cite{Wang_2012}. Most simply use past events to forecast future
events \cite{Chainey_2008,Kang_2017}, suggesting that methods
that are effective at forecasting crime could readily be generalized to an
increasing number of real-time spatiotemporal forecasting problems
\cite{taddy_autoregressive_2010}. However, users of
these methods often confront the question of which
 method to adopt and how to ensure optimal performance across a wide variety of settings.

In 2016 the National Institute of Justice (NIJ) announced the Real-Time
Crime Forecasting Competition to test which forecasting models could most
accurately predict out-of-sample crime hotspots in the City of Portland. This solicitation drew in a wide range of competitors. Teams were given five years of historical calls for service data from the Portland Police Bureau (PPB) and asked to submit predictions for the locations of the largest crime clusters in the subsequent weeks and months. 

Our team (``Team Kernel Glitches'') tied for first place in the large
organization category with wins across a range of categories. While our
solution performed equally well on frequent and sparse crime
forecasts and over short and long durations, it performed especially well,
compared to competitors and contemporary methods, at forecasting sparse events
over short durations. In describing our solution, we make the following
contributions: we propose a flexible, generic, and scalable spatiotemporal
forecasting model, casting the problem of spatiotemporal forecasting explicitly
as a supervised learning problem, while incorporating existing and highly
successful modeling approaches from the spatiotemporal statistics literature:
Gaussian processes, autoregressive terms, kernel smoothing, and self-exciting
point processes.  This supervised learning setup provides a coherent framework
for the time-consuming task of optimizing hyperparameters, while its modeling and inference scalability ensures that the model parameters themselves
can be learned quickly enough to enable real-time forecasting. This approach
achieves accuracy improvements well beyond those generated by existing
best-practices in crime prediction \cite{Chainey_2008,johnson_predictive_2009}. 

The rest of this paper is laid out as follows. Section \ref{section:model} describes our model. Section
\ref{section:competition} describes the details of the NIJ competition. Section
\ref{section:results} reports competition performance. Section
\ref{section:discussion} concludes with a discussion of implications for future
work on spatiotemporal prediction of crime and related phenomena.

\section{Our model}
\label{section:model}

\subsection{Background}
Previous methods for spatiotemporal forecasting of crime have either focused on highly flexible but relatively simple kernel density estimation techniques \citep{johnson_predictive_2009,gorr_early_2015}, where crime events are aggregated over time, smoothed over space, and used to  predict crime patterns in the subsequent time period, or more complex and model-based approaches \citep{Mohler_2011,rosser2016improving}. Recent work has demonstrated that Gaussian process modeling of crime data can produce highly accurate long-term forecasts by combining the benefits of nonparametric methods with the interpretability of additive methods \citep{flaxman2014general}. Subsequent work \citep{flaxman2015fast} has proposed that instead of specifying an additive kernel structure, it is is possible to learn it directly from the data, given enough data and a rich enough class of kernels. This assumes, however, that it is possible to perform inference with very large datasets, as the standard approach to Gaussian process inference requires matrix algebra to manipulate the multivariate Gaussian distribution in Eq.~\eqref{eq:mvn}, requiring $\mathcal{O}(N^3)$ time and $\mathcal{O}(N^2)$ storage. We therefore first present the hypothetical model we would use if computational constraints were not a concern,
then our actual model, which is an approximation to this model enabling application of this method to real-time rather than long-term forecasting problems. 

\subsection{Model specification}
Our hypothetical
model is a log-Gaussian Cox Process (LGCP). The LGCP
is a doubly stochastic point process model. 
Given an observation window $W$ in space-time,
we place a GP prior on the log-intensity $f(\bs)$ for any $\bs \in W$.
Let $N(\cdot)$ be a counting measure. 
For any space-time region $S \subset W$, $N(S)$ is a Poisson distributed 
random variable counting the number of points in 
$S$. Our hierarchical parameterization is as follows:
\begin{align}
	f &\sim \mathcal{GP}(\mu,k_{\theta}(\cdot,\cdot))  \nonumber\\
	N(S) | f &\sim \mbox{Poisson}\left(\int_S \exp(f(\bs)) d\bs \right) \label{eq:poisson}
\end{align}
We defer the specification of the mean $\mu$ and covariance kernel $k_{\theta}$ 
until later. For details on Gaussian processes see Appendix \ref{section:gps}.

Inference with the LGCP model is difficult because it is doubly intractable and existing approaches
\cite{moller1998log,brix2001spatiotemporal,cunningham2008fast,adams2009tractable,teh2011gaussian,diggle2013spatial}
are often limited to one dimension and small datasets.
\citet{lloyd2015variational} is a possible exception in that it points the way
to a scalable stochastic variational inference approach. 

To approximate this model we discretize. We specify a space-time grid partitioning $W$ into $N$ disjoints sets $S_i$, that
is
$W = \bigcup_{i=1}^N S_i$. 
As described below, this approach leads to a tractable model. Also, it is consistent with the design of the forecasting competition motivating our approach.
For simplicity, let each grid cell $S_i$ be of equal  volume $|S_i| = 1$.
The centroid of each grid cell is a latitude/longitude/timestamp triple $\bs_i = (x_i,y_i,t_i)$.
The underlying point pattern is then represented as aggregate counts $o_i = N(S_i)$ of the number of crimes per 
cell. Given the grid, the integral in Eq.~\eqref{eq:poisson} is approximated with a sum. When considering
the entire observation window $W$, the approximation takes the following form:
\begin{equation}
	\int_W \exp(f(\bs)) d\bs \approx \sum_{i=1}^N \exp(f(\bs_i)  )|S_i|  = \sum_{i=1}^N \exp(f(\bs_i))
	\label{eq:integral-approx}
\end{equation}
In a Poisson process, conditional on the intensity, the random variables $N(S_1)$ and $N(S_2)$ 
are independent for $S_1 \cap S_2 = \emptyset$.
Thus given the log-intensity $f$, each grid cell $S_i$ can be considered independently, so combining
Eqs.~\eqref{eq:poisson} and \eqref{eq:integral-approx} yields:
\begin{equation}
	o_i | f \sim \mbox{Poisson}\left(\exp(f(\bs_i)) \right), ~~\forall i = 1, \ldots, N
	\label{eq:poisson-likelihood}
\end{equation}
This produces an iid likelihood (observation model) over all cells $i$, yielding the so-called computational
grid approximation to the log-Gaussian Cox Process \cite{diggle2013spatial,flaxman2015fast}.

In the function-space view of GPs, inference is performed about the function $f$ directly.  Using the ``kernel trick'' \citep{scholkopf2002learning}, all calculations can be
carried out using a kernel $k_{\theta}$, evaluated at all pairs of  
$\bs_1, \ldots, \bs_N$. However, to do this requires storing and manipulating an $N \times
N$ covariance matrix $K$ at a cost of $\mathcal{O}(N^2)$ storage and
$\mathcal{O}(N^3)$ computation \citep{rasmussen2006gaussian}, which is
infeasible for large $N$.  

By contrast, the weight-space view of GPs \cite[Ch.~2]{rasmussen2006gaussian} requires an
explicit feature map $\phi(\bs) = k_{\theta}(\bs,\cdot) \in \mathcal{H}$ where
$\mathcal{H}$ is the Reproducing Kernel Hilbert Space corresponding to the
kernel $k_{\theta}$, with $\phi(\bs)^{\top}\phi(\bt) = k_{\theta}(\bs,\bt)$.
Instead of learning $f$ directly (function space), for finite dimensional
$\mathcal{H}$, a set of weights $\beta$ can be learned by considering
the vector $\phi(\bs)$ as a set of basis functions.
Thus we define $f(\bs) := \phi(\bs)^{\top} \beta$ and observe that the weight-space view is equivalent to a linear model with a particular set of basis
functions.  

In practice, the weight-space view is not computationally tractable in the case of
popular universal \cite{sriperumbudur2011universality} kernel choices like the  Gaussian or Mat\'ern kernel because the corresponding $\mathcal{H}$ is infinite dimensional.
Unlike infinite-dimensional universal kernels, kernels corresponding directly to finite-dimensional RKHS are limited in their representational capacity, e.g.~polynomial kernels of order $p$ only capture
$p$ moments of a distribution. A solution can be found, following recent trends in the literature
\cite{may2017kernel}, using finite-dimensional approximations to universal 
kernels in the form of the random Fourier feature expansion \cite{rahimi2007random} as described in Appendix \ref{section:scalable-kernel}. For any kernel,  this requires 
the selection of a dimension $d$ which determines the accuracy of the
approximation $\hat \phi \in
\mathcal{R}^{2d}$ where $\hat \phi(\bs)^{\top}\phi(\bt) \approx
\phi(\bs)^{\top}\phi(\bt) = k_{\theta}(\bs,\bt)$.  An example of our
approximation is illustrated in Figure \ref{fig:rff-compare} where the
Mat\'ern-5/2 kernel is approximated using various values of $d$. 

A finite dimensional $\hat \phi$ leads from the function-space view to the weight-space view
\cite[Ch.~2]{rasmussen2006gaussian}, \cite{milton2019spatial}. 
To make the connection explicit we define a kernel $\hat k_{\theta}(\bs,\bt) = \hat\phi(\bs)^{\top}\hat\phi(\bt)$. 
Define a matrix $\Phi$ for observations $\bs_1, \ldots, \bs_N$ with each row $\Phi_i = \hat\phi(\bs_i)^{\top}$.
The function-space view on
Gaussian process regression with covariance kernel $\hat k$ and a Gaussian likelihood is:
\begin{align}
	f &\sim \mathcal{GP}(\mu, \hat k_{\theta}(\cdot,\cdot)) \nonumber\\
	y | f, \bs_i &\sim \mathcal{N}(f(\bs_i),\sigma^2) \label{eq:gp-formulation}
\end{align}
Eq.~\eqref{eq:gp-formulation} is equivalent to Bayesian linear regression with $\beta \in \mathcal{R}^{2d}$ (where 
the term weight-space view comes from considering the parameter vector $\beta$ as ``weights'' to be learned):
\begin{align}
	\beta & \sim \mathcal{N}(0,I) \nonumber\\
	y | \beta, \bs_i, \Phi & \sim \mathcal{N}(\mu(\bs_i) + \Phi_i \beta, \sigma^2)
	\label{eq:linear-formulation}
\end{align}
For the present application, the data consists of count-valued observations, so we adopt a generalized linear modeling (GLM) framework
and replace the Gaussian likelihood in Eq.~\eqref{eq:linear-formulation} with the Poisson likelihood as in Eq.~\eqref{eq:poisson-likelihood}:
\begin{equation}
	o_i | \beta, \bs_i, \Phi \sim \mbox{Poisson}\left(\exp(\mu(\bs_i) + \Phi_i \beta)\right)
	\label{eq:glm}
\end{equation}

It remains to specify the function $\mu$. 
In the spatial statistics literature, a linear model using
spatially varying covariates is standard (e.g.~\cite{diggle2013spatial}),
while $\mu = 0$ is a common default choice in machine learning, though
recent work has questioned this approach \cite{bhatt2017improved}.
We consider a different approach, based on prior work that has shown that
using historical crime rates can be very effective in crime forecasting. 
Expanding upon prior KDE-forecasting methods that search a limited number of possible values and in line with the supervised learning framework discussed above, $\mu$ is parameterized as follows for $\bs = (x,y,t)$:
\begin{align}
	\mu(\bs) = \sum_{j=1}^{p} \gamma_j \mbox{KDE}_{\lambda,j}(x,y,t)
\end{align}
where there are $p$ autoregressive lagged terms, each representing a spatial KDE for a 
given time period in the past and regression coefficients $\gamma_j$ are to be learned.  $\mbox{KDE}_{\lambda,j}(x,y,t)$ is the kernel
density estimator at location $(x,y,t)$
using a spatial Gaussian kernel $\kappa_\lambda$ with lengthscale $\lambda$:
\begin{equation}
	\mbox{KDE}_{\lambda,j}(x,y,t) = \sum_{\{t_i ~|~ t - j \cdot D ~<~ t_i ~\leq~ t -
(j-1) \cdot D \}} \kappa_\lambda((x,y),(x_i,y_i))
	\label{eq:kde}
\end{equation}
where $D$ is the size of the temporal window in days.

Given the potential for a large number of parameters $\beta$ (the more
random frequencies $d$ we choose for the random Fourier feature expansion, the
better our approximation), the use of $\ell_1$ and $\ell_2$
regularization (as in the popular elastic net
\cite{zou2005regularization}) provides a useful simplification.

Finally, our objective is to maximize the penalized log-likelihood of the Poisson distribution. Simplifying and dropping constant terms yields the following objective, with parameters $\beta$ and $\gamma$ and
regularization hyperparameters $a$ and $b$:
\begin{align}
	\sum_{i=1}^N & \left[  o_i \left( \sum_{j=1}^p \gamma_j \mbox{KDE}_{\lambda,j}(x_i,y_i,t_i) + \Phi_i \beta \right) - 
	e^{\sum_{j=1}^p \gamma_j \mbox{KDE}_{\lambda,j}(x_i,y_i,t_i)  + \Phi_i \beta} \right] \label{eq:objective}\\
	& - a(\|\beta\|_1 + \|\gamma\|_1) - b (\|\beta\|_2^2 + \|\gamma\|_2^2) \nonumber
\end{align}

\subsection{Inference}
We learn the parameters $\beta$ and $\gamma$ by maximizing the objective in Eq.~\eqref{eq:objective}
using gradient ascent.  The random Fourier feature approximation combined with
linear regression leads to immediate speed-ups and memory savings: whereas full
GP regression is $\mathcal{O}(N^3)$ time and $\mathcal{O}(N^2)$ storage,
calculating the random features for $\Phi$ is $\mathcal{O}(Nd)$ for both time
and storage. Given a fixed design matrix $\Phi$, ordinary linear regression requires
calculating $\Phi^{\top}\Phi$ which is $\mathcal{O}(Nd^2)$ time and
$\mathcal{O}(d^2)$ storage.  Depending on how the lasso and ridge penalties are
implemented, penalized linear regression can be very efficient, e.g.~cyclical
coordinate descent takes $\mathcal{O}(Nd)$ time for each update of all of the parameters \cite{friedman2010regularization}. The important point is that the overall running time is linear in $N$ rather than cubic, a significant savings in time.
This approach is competitive with standard approaches to scalable inference in
the spatial statistics literature
\cite{sun2012geostatistics,milton2019spatial}. 

During competition, we performed optimization using the large-scale machine
learning package Vowpal Wabbit (\url{http://hunch.net/~vw}). Vowpal Wabbit
employs feature hashing \cite{weinberger2009feature} and online learning, which
is even faster than the standard $\mathcal{O}(Nd^2)$ approach to linear
regression, allowing it to scale up to handle huge datasets. We fit the
training dataset using default settings for the learning algorithm (a variant
of online gradient descent), with at most 200 training passes (epochs) through
the dataset. As a stopping criterion for convergence was applied, running times
did not directly vary with dataset size. For any given set of hyperparameters
(except the regularization parameters), a new dataset was produced and saved to
disk, and then this model was fit across the full path of regularization
parameters by repeatedly running Vowpal Wabbit. The entire process of dataset
creation and multiple calls to Vowpal Wabbit usually took about half an hour,
even with datasets as large as $N \approx 300k$ (1 week time horizon). All of
our computation was carried out in a parallel cluster computing environment,
with 8 Dell PowerEdge R630 nodes. Each node consisted of 2 $\times$ Intel Xeon
E5-2690 v4 2.6 GHz, 14 Core CPUs, and 256 GB memory. 

After fitting the model, we made predictions in the form of counts
for the test data, and then calculate PEI for each year of data, where PEI is a
forecasting accuracy metric used in the crime forecasting literature.
To learn the hyperparameters we maximize average PEI.  The hyperparameters
related to our model are as follows: the number of random features $d$ in our
feature expansion, the number of lags $p$, the size of the temporal window $D$,
the spatial lengthscale for KDE $\lambda$ (with a Gaussian kernel), the
lengthscale $\theta$ of the covariance kernel $k_{\theta}$ (we used a Mat\'ern-5/2 kernel, a standard choice
in spatial statistics \cite{guttorp2005whittle}), and
the amount of $\ell_1$ and $\ell_2$ regularization $a$ and $b$. In addition, there are competition-related hyperparameters that are learned, including: cell size, shape, grid rotation, and forecast area. We crossvalidated over a very large grid of hyperparameters, considering
a range of values for each parameter and every possible combination of these values.
As an alternative method to  further explore the entire space of hyperparameter choices, we separately performed hyperparameter search using sequential Bayesian Optimization
\cite{ohagan1992some,snoek2012practical,HenOsbGirRSPA2015}. Having run both searches, we combined the results and chose the best sets of hyperparameters based on crossvalidated average PEI. Additional details are given in Appendix \ref{appendix:hypers}.

\subsection{Relationship with prior work}
Supervised learning methods are widely used within non-spatiotemporal applications. However, they are less commonly used within the applied spatial \cite{heaton2017case}, time series \cite{makridakis2018statistical}, and crime forecasting domains. In crime forecasting, KDE-based forecasting approaches remain the most common forecasting techniques used \cite{gorr_short-term_2003,gorr2009forecast,chainey_utility_2008,Caplan_2011,berk2018fairness}. While small numbers of parameters may be user-selected and modified, these methods are commonly implemented absent any framework for maximizing the objective function of forecasting accuracy. Instead, practitioners modify parameters on an ad hoc basis, assuming that the resulting forecasts are a reasonable implementation of KDE methods. For a recent exception to this approach, see \citet{rosser2016improving}. 

When prior work has sought to improve upon the performance of these less-than-optimized KDE forecasts, the principle area of focus has not been on scalable hyperparameter optimization, but instead on implementing model-based characterizations of underlying crime intensities. Some work has focused on modeling spatial and temporal range of crime decays \cite{johnson_predictive_2009}, but the Hawkes process has recently been the focus of significant attention in the crime forecasting literature \cite{ogata1988statistical,moller2005perfect,Mohler_2011,mohler2013modeling,mohler2014marked,rosser2016improving,loeffler2018gun}. Both approaches seek to avoid a common feature of prior KDE methods which implicitly weight all prior events as equally informative with no attention to recency. However, the question of how to identify the optimal spatial and temporal range of crime decay is also not entirely addressed in these contributions.  

The logic of our approach is that it combines state-of-the-art nonparametric spatiotemporal methods (Gaussian process regression), which fundamentally encode an assumption of spatial and temporal autocorrelation, with the most long-standing and widely used crime forecasting method (KDE surfaces) by
defining sets of features for each. 
By placing these two sets of features into
a penalized supervised learning framework for forecasting the intensity, and considering
a large set of hyperparameters and training data, we hope to  combine the
benefits of nonparametric modeling, principally accuracy in the absence of a
known best model, with the benefits of parametric modeling, principally model
simplicity, to obtain good predictive performance on unseen data. For a discussion of the similarities of optimized KDE features and so-called ``Hawkes features'', see  Appendix \ref{appendix:hawkes-features}.

\section{The competition}
\label{section:competition}
The goal of the NIJ Real-Time Crime Forecasting Competition was to forecast hotspots for several categories of calls for service to the Portland
Police Bureau (PPB) in Portland, Oregon. Contestants submitted forecasts on (or
before) February 28, 2017 for various time horizons starting on March 1, 2017
and extending as late as May 31, 2017. The hotspot predictions were scored on two metrics related to their accuracy.  Contest rules required that
contestants predict which of the 62,500 - 360,000 square foot
cells within PPB's 147.71 square mile service area would
have the highest number of calls for service, with the total forecast area
being no smaller than 0.25 square miles and no larger than 0.75 square miles,
equivalent to forecasting 175--525 city blocks out of a total of 103,397
blocks. Prizes were given out for five different cumulative forecast periods (1 week, 2
weeks, 4 weeks, 8 weeks, 12 weeks), four different crime categories (burglary,
street crime, theft of auto, all calls for service), and two different
accuracy metrics.

\begin{figure}[ht!]
	\makebox[\textwidth]{
	\includegraphics[width=1\linewidth]{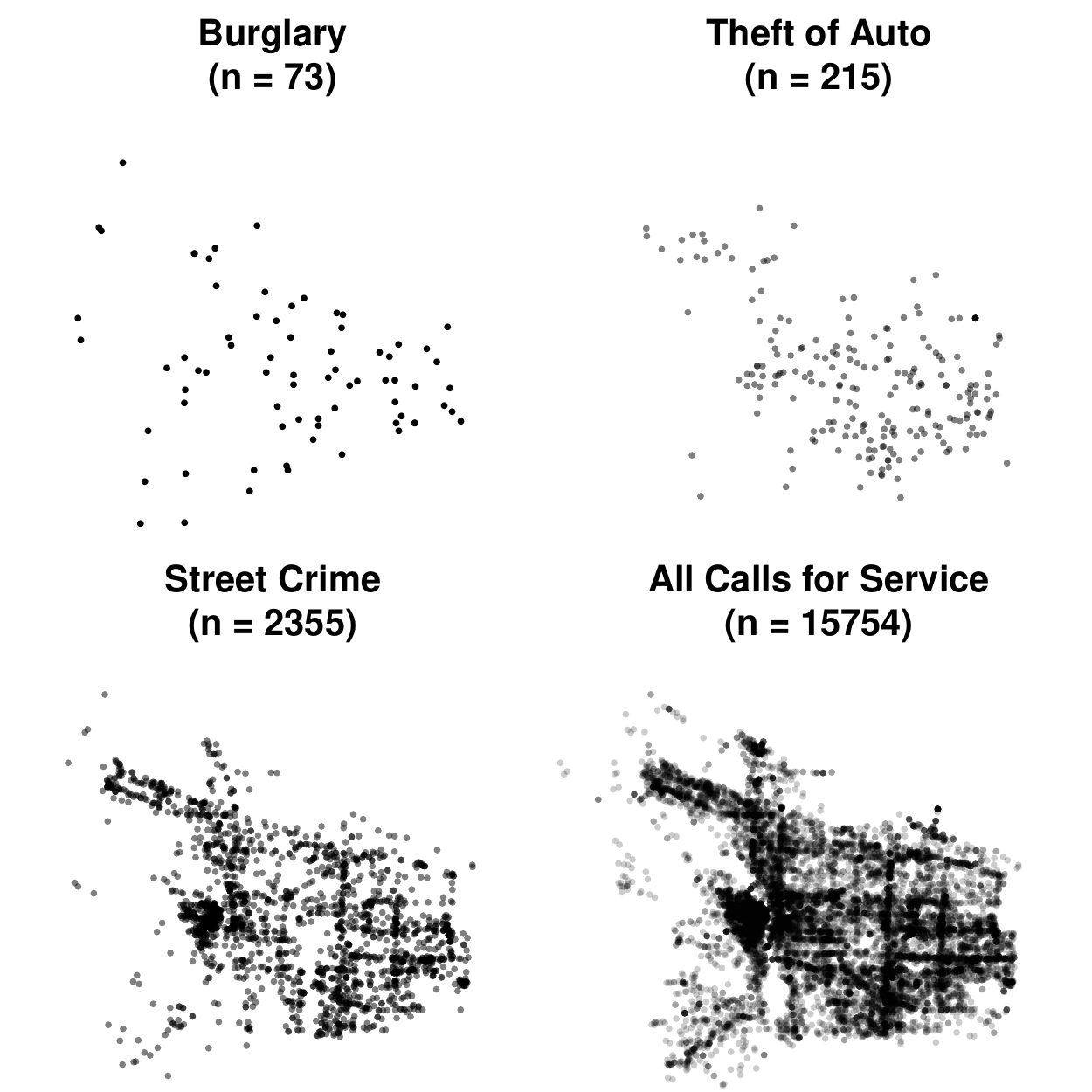}
	}
	\caption{The competition focused on four categories of crimes, ranging from the very abundant (All Calls for Service) to the very sparse (burglaries). Shown here are the locations of reported crimes in February 2016.}
	\label{fig:crimes-feb-2016}
\end{figure}

\subsection{Data and Setting}
The NIJ Real-Time Crime Forecasting dataset consists of 958,499
calls for service records from the Portland Police Bureau (PPB), representing calls to Portland's 911 system requesting police assistance from March 1st, 2012 through February 28th, 2017. As shown in
Figure \ref{fig:crimes-feb-2016}, the four categories of crime, which themselves
varied in the degree of internal heterogeneity, included burglary (burglary and
prowling), street crime (ranging from disturbance and threats up to armed
robbery and assault with a Firearm), theft of auto, and all calls for service.

\subsection{Metrics}

The simplest metric for evaluating the accuracy of crime forecasts is the ``hit rate'' \cite{chainey_utility_2008}\footnote{It is also known as sensitivity in the statistics literature. See \citet{adepeju_novel_2016} for a recent discussion of alternative evaluation metrics for crime forecasting.}:
$$ \mbox{Hit rate} = \frac{n}{T}$$
where $n$ is the number of crimes predicted and $T$ is the total number of crimes in that period in that area. Performance on this metric depends critically on the size of the forecasted area in addition to underlying crime densities and forecasting quality. In the case of the NIJ competition, this coverage area was between 0.2\% and 0.5\% of the City of Portland. 

The NIJ competition focused on two alternatives metrics \cite{chainey_utility_2008, hunt2016crime}, with
a goal of allowing for a comparison of hit rates across
forecasts using different coverage areas. The first metric, the prediction accuracy index (PAI) \cite{chainey_utility_2008}, is the ratio of the hit rate to the fraction of area covered:
$$ \mbox{PAI} = \frac{\frac{n}{T}}{\frac{a}{A}}.$$
This metric directly incorporates the trade-off 
between hit rate and coverage, as in an ROC curve,
into the score weighting.

The second metric, the prediction efficiency index (PEI), is the ratio of PAI to the hypothetically maximum PAI that
could have been obtained using the chosen coverage area and discretization of space. Since the forecasting area is the same
in both the actual and hypothetical maximum cases, this reduces to:
$$ \mbox{PEI} = \frac{n\hfill}{n^{*}}. $$
where \(n\) is the number of crimes occurring in predicted hotspots,
and \(n^{*}\) is the maximum number of crimes that could have been captured for the forecasted area. 

While optimizing either metric will produce similar results some of the time, optimizing for PEI incurs a PAI penalty proportional to the
marginal change in forecasted area divided by the marginal change in correctly forecasted crimes. Therefore, the optimal cell selection for maximizing PEI will often fail to maximize PAI. For the competition, we maximized the PEI metric. (For a result making the opposite choice, see \citet{mohler2018rotational}.)

\subsection{Data for training and hyperparameter selection}
For a given spatial grid size, we restricted our temporal windows to match the corresponding forecasting window. For example, for a one week forecasting window, the training data is aggregated to the weekly level. The training period consisted of each prior year's aggregated counts excluding the corresponding time period being forecasted. This excluded period formed the validation period. We then created a single dataset using data from the union of all of the training and validation periods. Using this dataset, we forecasted hotspot maps for the five different validation periods, corresponding to the five different years of pre-2017 data available and calculated PEI for each. The average of this heldout PEI was then maximized to select the hyperparameters of the model.

\section{NIJ Challenge Results}
\label{section:results}

In this section we describe the performance of our method according to the
scoring metrics of the NIJ challenge, assess its robustness, and investigate what features of the model contributed to its out-of-sample performance.

There were a total of 40 prizes awarded, one for each of the highest PEI and PAI scores in each crime category and forecasting window. Our team won a total of 9 prizes in the ``Large Business'' competition. As we focused on maximizing the forecasting performance on the out-of-sample PEI metric, most of our winning entries were in this category:
all calls for service (1 week, 1 month, 3 months), burglary (1 week, 2 weeks), street crime (2 weeks), and theft of auto (1 week). In addition, we also had winning PAI entries for burglary (1 week and 2 weeks).

At the heart of our model was a hyperparameter search strategy, in which final models were selected from the union of all models explored by an exhaustive grid search coupled with a Bayesian Optimization designed to optimize forecasting accuracy. In practice, there were no consistently chosen hyperparameter values: the grid cells were
sometimes small squares $250 \mbox{ft} \times 250 \mbox{ft} $ (the minimum
area) or large squares $600\mbox{ft} \times 600\mbox{ft}$ (the maximum area) or
large rectangles $800\mbox{ft} \times 450\mbox{ft}$ (also the maximum area).
The coverage fraction ranged from the minimum (0.25 sq miles) to the maximum
(0.75 sq miles). The lengthscales for space and time were highly varied,
as were the number of KDE lags and the KDE bandwidth. The number of random Fourier features went as low as $d=5$, which means that the
surface was a very crude approximation to a Gaussian process consisting of the
sum of $10$ random sine and cosine functions, to as high as $362$, a much better
approximation. In a minority of cases, no $\ell_1$ or $\ell_2$ regularization was
needed, but most final models used at least some $\ell_2$ regularization. In a minority of cases (4 out of 20) the best hyperparameters turned out to be those
found by Bayesian Optimization, while in all other cases, the 
best hyperparameters were those found by grid search. (See Table \ref{table:hypers}  for details.)
The lack of overlap in optimal hyperparameter selection across competition
categories both reinforces the importance of supervised learning optimization
for forecasting accuracy and raises the question of whether other, possibly
more uniform, hyperparameter choices might also exist.

We examine the distribution of all PEI values obtained in our grid search for
each category/forecast window separately. 
For the 1 week  theft of auto and burglary categories, 41\% and
44\% (respectively) of the possible hyperparameter combinations gave PEI scores
of 0. This is strong evidence for the importance of an exhaustive hyperparameter
search, at least for these sparse events. To further quantify this numerically, 
we calculate the z-score of the maximum PEI for the distribution of PEIs for
each category/forecast window. 
Our winning theft of auto 1 week entry had a PEI z-score of 21,
and our winning burglary entries had z-scores of 12.4 (1 week) and 11 (2
weeks), all results which are consistent with the idea that good forecasting
accuracy requires an exhaustive hyperparameter search. The distributions
for more abundant crime types did not yield such extreme z-scores:
in the All Calls for Service category, the z-scores of
the maximum PEIs ranged from 2.5 to 4.0. In the street crimes category the
z-scores of the maximum PEIs ranged from 2.8 to 5.6. Thus for more abundant
crime types a range of hyperparameters could produce similar results. 

\begin{figure}[ht]
	\centering
	\includegraphics[width=.9\textwidth]{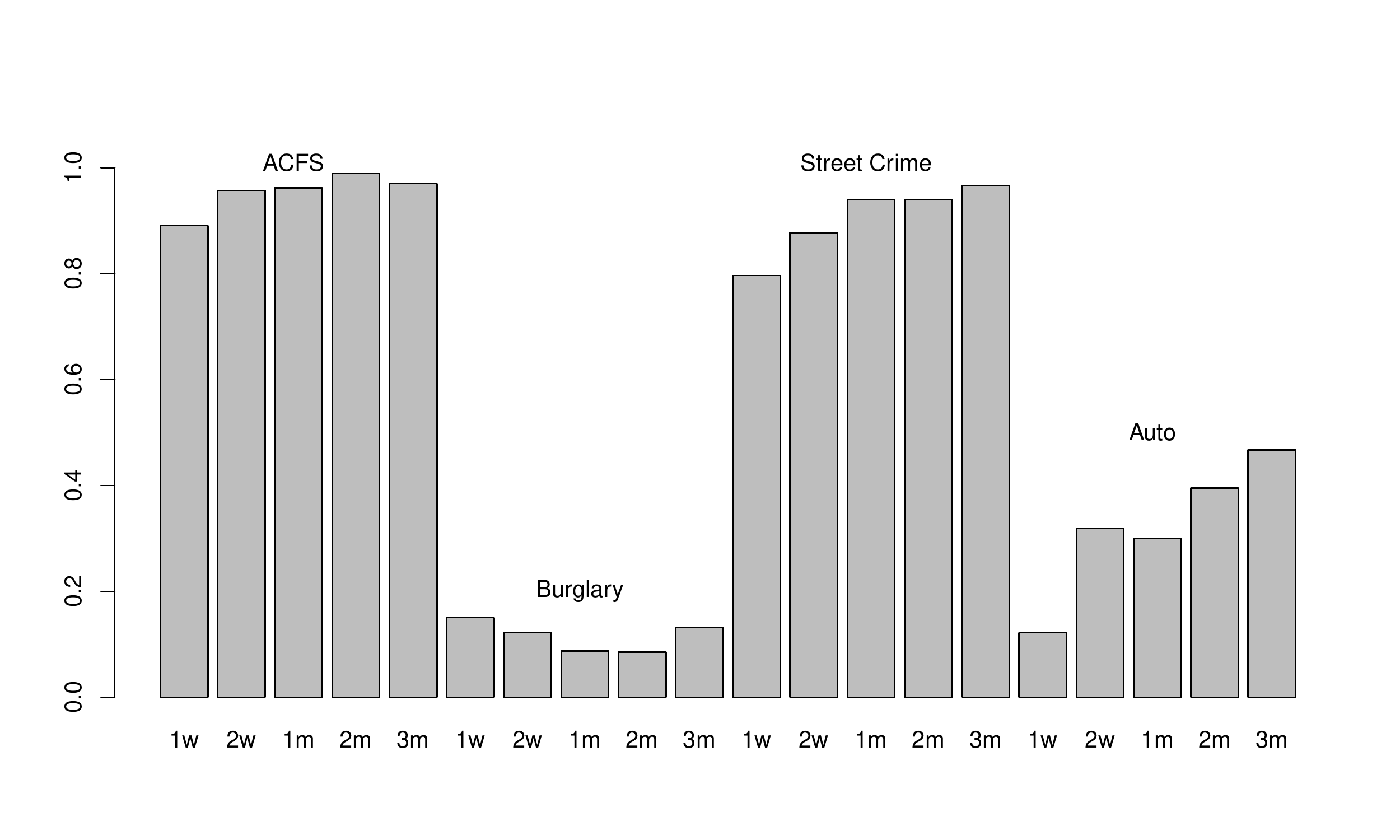}
	\caption{Competition Maximum PEI Performance Among All Competitors. Each column represents the highest level of forecasting accuracy achieved across all competitors for a particular combination of crime type and forecasting period. From left to right all calls for service, burglary, street crime, and theft of auto for 1 week, 2 weeks, 1 month, 2 months, and three months (also left to right).}
    \label{fig:maxpei}
\end{figure}
A final question concerning the competition is the maximum achievable level of forecasting accuracy. As shown in Figure \ref{fig:maxpei}, which depicts the maximum achieved PEI for all competitors, for high volume crimes, such as all calls
for service, even a week's worth of data is sufficient to achieve very high PEI
scores (nearly 0.9) of the theoretical limit (1) for a one week prediction.
Extending the cumulative forecast period leads to further improvements in forecasting
accuracy, plateauing at 97\%. Sizable sub-categories, such as street crimes,
share this basic trajectory as well, suggesting that for high volume crimes over both short and medium-term horizons near limit and unity performance can be expected. 
For some sparse crimes, such as theft of auto, despite lower starting values, similar improvements in
predictive accuracy can be seen as the forecasting windows are expanded, even if these improvements are not strictly monotonically increasing. Whether a longer horizon would lead to further improvements is unknown.
However, for other sparse crimes, such as burglary, adding additional weeks of
data to the forecast period does little to improve maximum achieved forecast accuracy. Reinforcing the idea that crime forecasting is not a single problem but several, only some of which are more accurately solved through the addition of more data.

\subsection{Investigating method performance}
\begin{figure}[ht!]
	\centerline{
		\includegraphics[width=.9\textwidth]{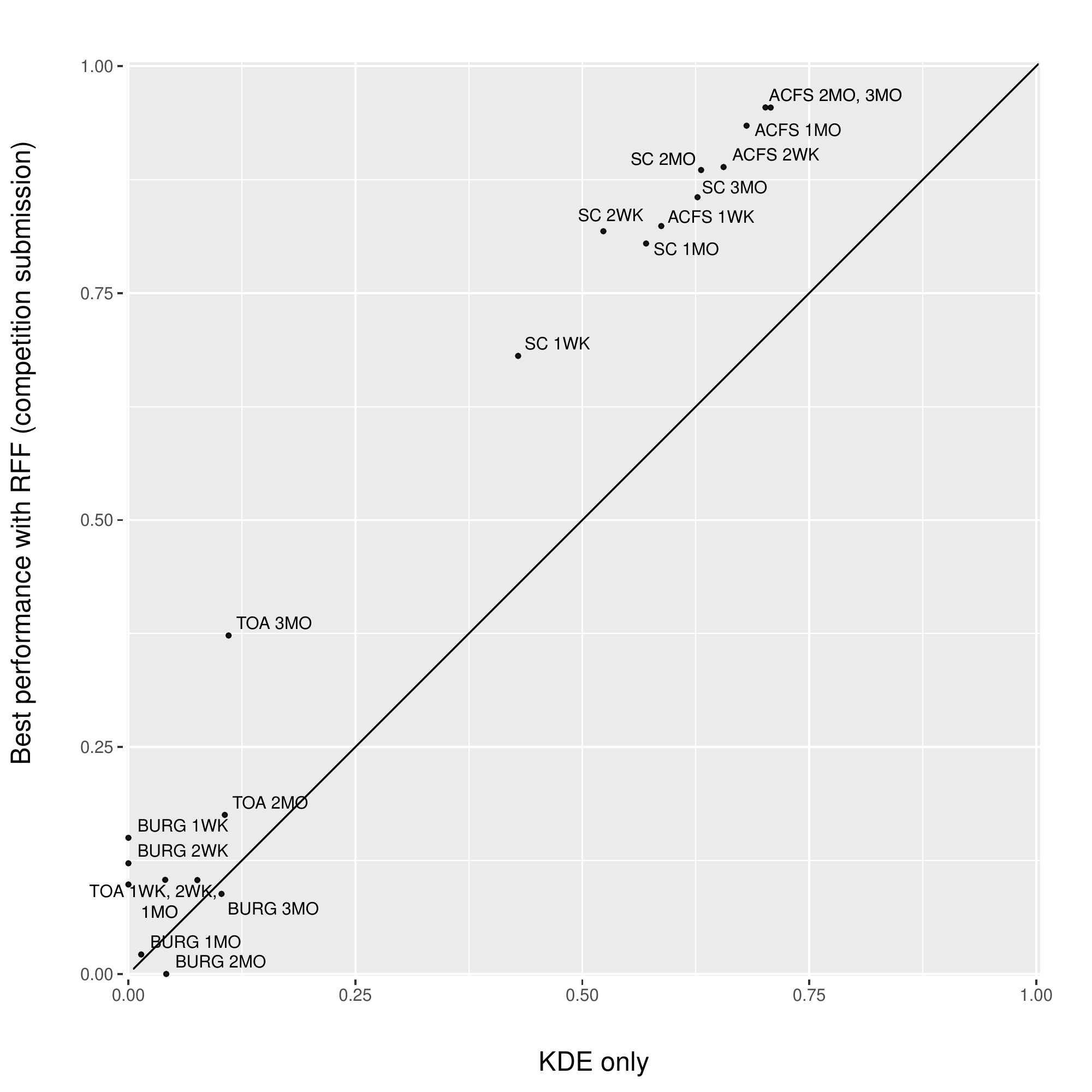}}
\caption{KDE Baseline (x-axis) compared to Full Model (y-axis). The Full Model out-performed the KDE Baseline Model in 18 out of 20 forecast problems. The average out-of-sample performance improvement of the Full Model over the KDE Baseline Model was 0.16 on the PEI scoring metric. BURG = burglary, SC = street crime, TOA = theft of auto, ACFS = all calls for service. }
\label{fig:sub1}
\end{figure}

\begin{figure}
\centering
\includegraphics[width=1.0\textwidth]{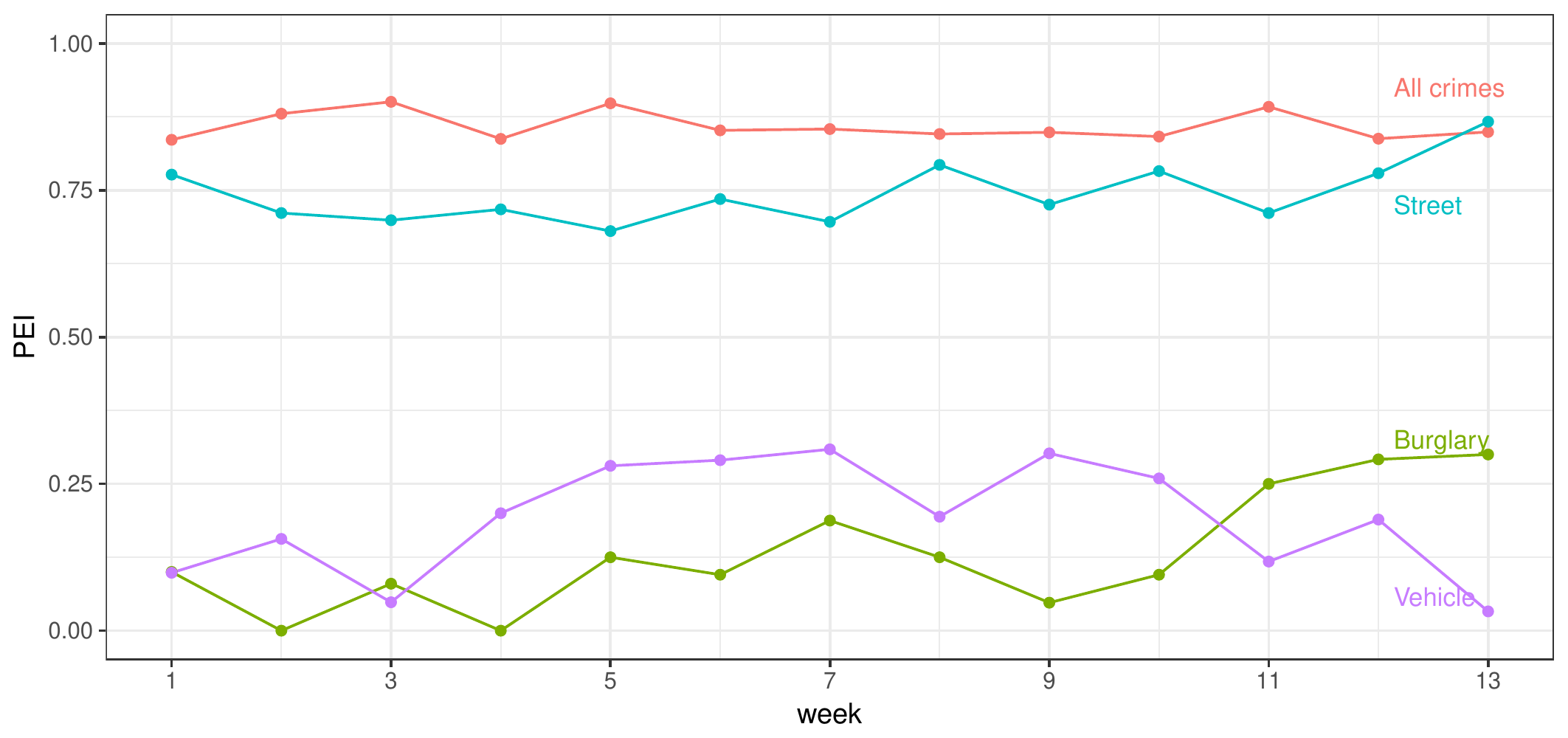}
\caption{Rolling Forecast. The 13 week competition period (March through May 2017) was split into 13 one-week forecast periods and a one-week rolling forward prediction was made for each week using the competition model trained to predict only the first out-of-sample week.}
\label{fig:Rolling_Forecast}
\end{figure}

As discussed in Section \ref{section:intro}, many crime forecasting implementations rely on 
KDE-type approaches. As our model included lagged KDE terms, we expected to always perform as well as a KDE-type baseline. As a post-competition check, we fit a model with just one KDE lag, corresponding
to a KDE-type baseline, and fixed parameters according to common practice \cite{chainey2005gis} and found that our model was better than this baseline 90\% of the time (18 cases out of 20) on the true out-of-sample forecasted data with an average absolute improvement of 0.16 for the PEI scoring metric. The improvements were most notable for sparse crimes
and short time-horizons, as the baseline model often identified no correct theft of auto or burglary hotspots (Figure \ref{fig:sub1}). Interestingly, for the two forecasts for which simple KDE outperformed our model (e.g., burglary 2m and 3m), hyperparameters for the model were selected using Bayesian Optimization rather than grid search, suggesting that BO will not always give the optimal set of parameters.
Comparing the full method, which combines lagged KDE terms and a Gaussian process surface, to a model without the Gaussian process surface, the full model gave better PEI results 75\% of the time. The average absolute improvement was 0.05. Thus, although the full model is  an improvement, 
the improvements are not as dramatic as going from a simple KDE to a lagged KDE model with kernels optimized for forecasting accuracy. This result suggests that for many models, especially ones predicting sparse events (as depicted in Figure \ref{fig:Rolling_Forecast}), the routine variation in performance is sufficient to swamp the benefits of using Gaussian process surfaces or other complex methods. Instead, considerable portions of achievable performance improvements can be realized by optimizing the parameters of simpler methods, such as lagged KDEs.

 \citet{Rosser_predictive_2016} recently demonstrated that due to geocoding, non-cardinal land use, and other related factors, a non-standard alignment could improve predictive accuracy in crime forecasting. In the present application, we explored altering the rotation of the entire tessellation and the dimensions of the cell rectangles. With improved performance of only 0.029 on the PEI scoring metric for a freely-rotated model when compared to the best performing non-rotated model, rotation does not appear to be a major contributor to overall performance. However, certain crime categories and forecast windows can be observed to benefit more substantially. A similar result can be observed for altering cell dimensions, which only improves overall performance on the PEI scoring metric by 0.019 when compared to a conventionally used 600x600 ft rectangle. (See Figures \ref{fig:sub2}--\ref{fig:sub4} for more details.) These results parallel previous findings that showed the limited return on the inclusion of non-auto-regressive information \citep{Wang_2012,gerber2014predicting}.

The sparseness of several of the forecasted incidents and recent findings on lack of robustness of forecasting models \cite{rosser2016improving} suggests that it is worthwhile to examine the stability of the model's performance over multiple periods. To accomplish this, the 13 week competition period (March through May 2017) was split into 13 one-week forecast periods and a one-week rolling forward prediction was made for each week. The resulting predictions, as seen in Figure~\ref{fig:Rolling_Forecast}, manifest variability consistent with the stochastic events being predicted. However, these rolling-forward predictions provide little evidence of over-fitting to the first out-of-sample time period, even for the sparsest of incidents. They instead suggest, at least for settings like the competition, that the short-term accuracy improvements are robust and stable. 

\begin{figure}[h]
	\centerline{
	\includegraphics[width=.49\textwidth]{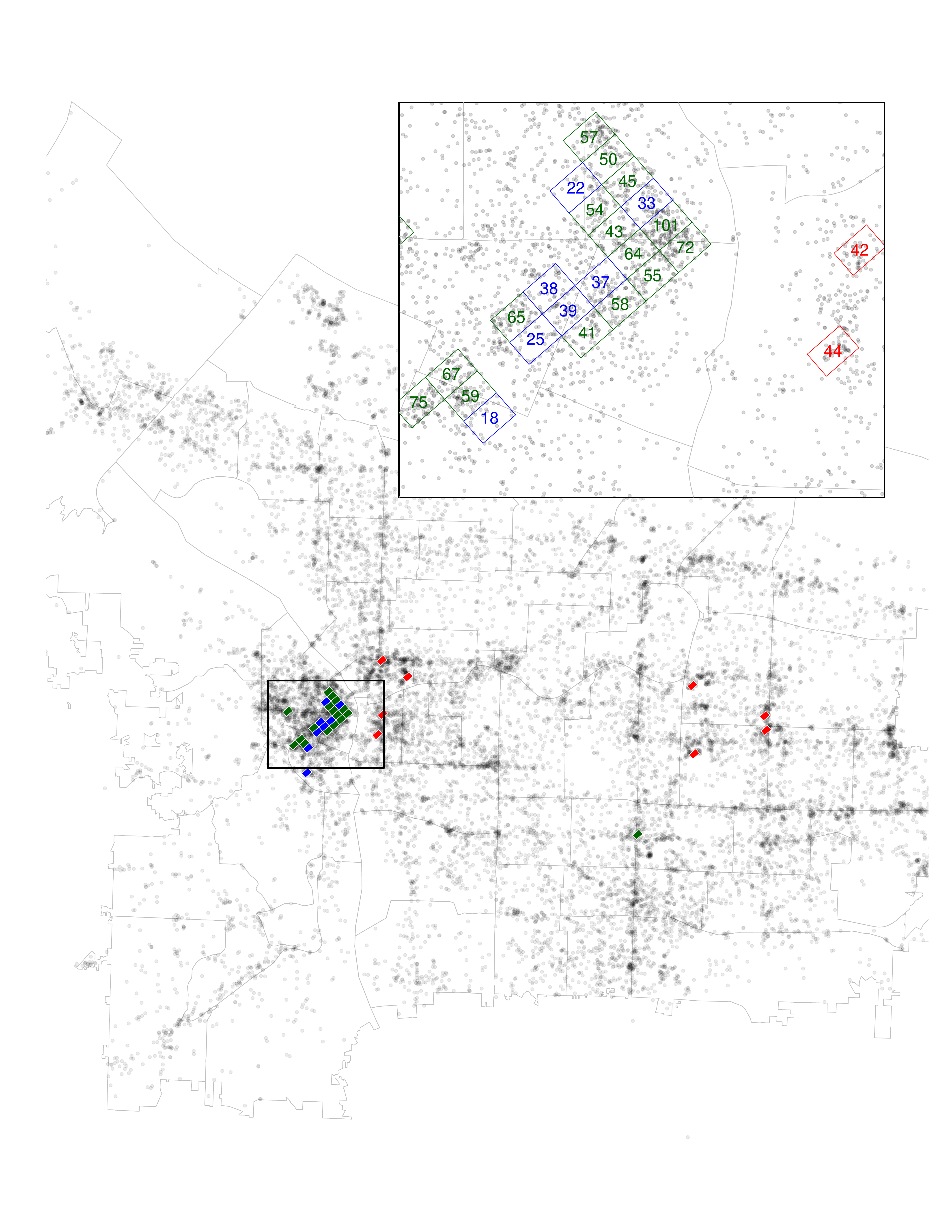}
	\includegraphics[width=.5\textwidth]{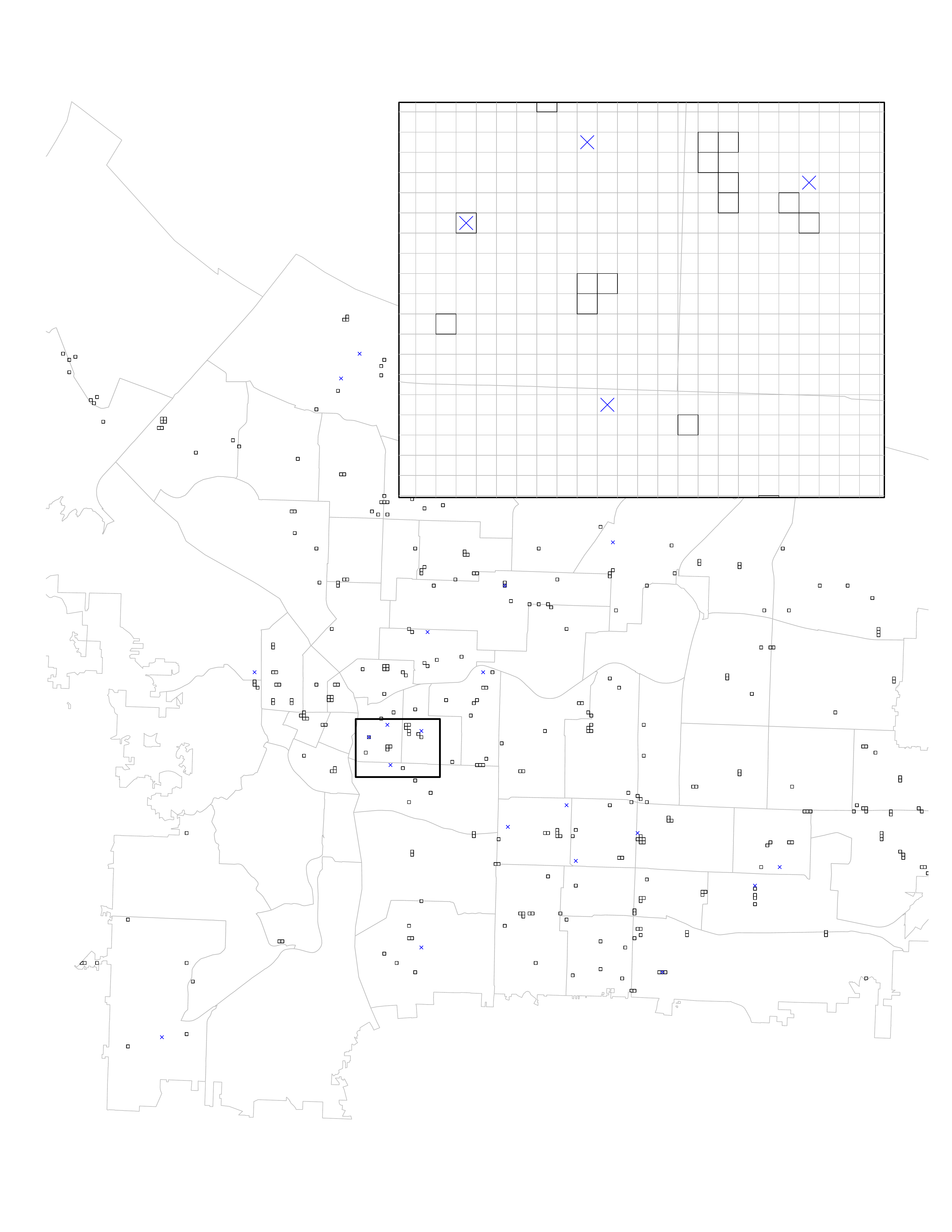}
	}
    \caption{Left: all calls for service 1 month. Correctly forecast polygons are in green. ``False negative'' polygons are in red. ``False positive'' polygons are depicted in blue. Crimes are black dots.
	Right: burglary 1 week. Forecasted burglary cells are depicted with boxes. Actual burglaries are depicted by blue x's. Boxed x's indicate a successful prediction. Empty boxes indicate a ``false positive'' prediction.}
    \label{fig:acfs}
\end{figure}

Alongside sub-model component performance and model stability, a final area of interest is method error. Figure \ref{fig:acfs} (left) shows the actual performance of the full forecasting model for a high volume crime category (ACFS) and a middle-range forecasting period (1 month). Polygons that were correctly forecast as the highest possible crime count polygons are in green. Polygons incorrectly forecast to not be hotspots are in red. And polygons that were incorrectly predicted to be the highest possible crime count polygons are depicted in blue. Crimes are black dots. The largest single cluster of hotspots for all calls for service can be seen downtown. However, the model slightly over-invested in this section of Portland. As can be seen in the inset, hotspots just across the Willamette River had more crimes reported over the relevant forecast window. In practice, most of these misses were relatively small, with ``false negatives'' only slightly ``hotter'' than the corresponding ``false positive'' cells (e.g., 44 crimes in a FN cell versus 39 crimes in a FP cell). 

Figure \ref{fig:acfs} (right) show the actual performance of the model for a sparse crime category (burglary) and a short-range forecasting period (1 week). Forecasted burglary cells are depicted with boxes and actual burglaries are depicted by blue x's. Boxed x's indicate a successful prediction while empty boxes indicate a ``false positive'' prediction. The absence of large-scale clustering is quite visible in both the dispersion of the burglaries throughout Portland and in the similarly dispersed allocation of predictions. As can be seen in the inset, a successful prediction was accompanied by several near misses in the vicinity, including one near-miss off by only a single cell. Predicting sparse crimes, while more difficult than predicting concentrated crimes, is still achievable and with accuracy levels not previously seen with other conventional forecasting methods.

\section{Discussion}
\label{section:discussion}

Real-time spatiotemporal forecasting is an area of increasing interest. Yet many common approaches, such as kernel-smoothing based on fixed bandwidths and cell sizes, can be quite limited in their out-of-the-box accuracy, especially for sparse events. Past work \cite{johnson_predictive_2009} has reported 1-week burglary forecasting accuracy of 10\% at 1.3\% of coverage area and 25\% of burglaries at 5\% of coverage area using near-repeat models with baseline KDE models producing 1-week accuracy of 10\% at 2\% coverage and 25\% at 6.5\% coverage. \citet{Mohler_2011} report 5\% accuracy for daily predictions at comparable coverage levels.  By comparison, using the described methods, median 1-week burglary accuracy of 10\% was achieved with a coverage area of 0.5\% and 50\% of the time 25\% forecasting accuracy was achieved at 0.5\% coverage. 

These results build upon prior work exploring parameter tuning \cite{chainey_examining_2013,rosser2016improving} and reinforce three points. First, it appears that simple but well-tuned models incorporating lagged kernel smoothing can achieve many of the benefits commonly associated with more complex methods. This conclusion stems from the recognition that parameter optimization, particularly in the case of kernel smoothing, is a re-weighting of different spatiotemporal portions of an auto-regressive process for forecasting accuracy. Second, the poor performance of conventional kernel estimators with parameters set based on rules-of-thumb, suggests that many existing crime forecasting implementations are not as accurate as they could be. Third, while some parameters are more important than others, no one parameter is universally better and as such, supervised learning will likely be a continuing feature of spatiotemporal crime forecasting. 

While the results reported here suggest that forecasting the hottest high volume crime hotspots can be done with great accuracy using a variety of techniques, the same cannot be said for sparse events, at least not yet. This leaves as an open question whether rare crime events are intrinsically harder to forecast due to random error or are simply harder because of insufficient training data. The fact that some rare crime forecasts saw no improvement in forecasting accuracy despite the addition of more training data and larger cumulative forecasting windows could be considered suggestive evidence that there may be a signal limit for this type of event. However, refitting our models in other settings would shed further light on this question, as would the inclusion of additional predictors. For example, $\mu$ based on the kernel density estimates of other  types of crimes, inspired by criminology research on 
``leading indicators'' of crime \cite{cohen2007leading}. 

Another question not answered by these results is why this method's performance was not more uniform. One possible answer is that the methods described in this paper simply do a better job at forecasting certain types of events over certain forecasting windows. Another possibility is that incomplete grid-search of hyperparamaters during competition led to the use of sub-optimal parameters for certain forecasting sub-tasks. A final possibility is that the close performance of competitors, on at least some forecasting tasks, achieved near limit forecasting performance using known methods and data. In future work in other settings, these possibilities could more readily be teased out. 

Pending completion of this research, the absolute performance of different methods in this competition also raises the policy question of what is an acceptable level of accuracy for any crime forecasting method to be used? 

In recent years, crime forecasting tools have been a supplement or replacement for traditional crime analysis \cite{mohler2015randomized}, with applications to police deployment, enforcement actions targeted at particular individuals or places  \cite{lum2016predict,perry_predictive_2013}, as well as non-enforcement notification strategies  \cite{groff2019using}. These applications, especially those involving law enforcement activity, have elevated concerns about fairness in criminal justice decision-making, leading to a vigorous debate about definitions of algorithmic fairness \cite{berk2018fairness,corbett2017algorithmic,mitchell2018prediction}. While fairness is an important debate, we have focused instead on accuracy, as this is a necessary precondition to considerations of fairness \cite{dressel2018accuracy,rudin2018optimized}. As the results of our research suggest, opportunities for large gains in accuracy exist through the use of standard machine learning frameworks and spatial statistical methods.

\section*{Acknowledgments}
Special thanks to our systems administrators: Tony Vo (University of Pennsylvania) and Stuart McRobert (Oxford).

\clearpage
\bibliographystyle{imsart-nameyear}
\bibliography{nij_bibliography.bib}

\newpage
\appendix
\setcounter{figure}{0}    
\setcounter{table}{0}
\renewcommand{\thetable}{A\arabic{table}}
\renewcommand{\thefigure}{A\arabic{figure}}
	\section{Scalable Gaussian processes}
\subsection{Gaussian processes}
\label{section:gps}
Following \citet{cressie2011}, we use Gaussian processes as the fundamental modeling approach for spatiotemporal data. In the particular case of point pattern data such as crime events, we follow \citet{diggle2013spatial} in considering the log-Gaussian Cox Process. 

A Gaussian process is a stochastic model which can be used as a nonparametric prior over functions $f$. See \citet{rasmussen2006gaussian} for a comprehensive introduction. $f$ is defined on some index set $\mathcal{X}$ and for our purposes we will assume that $f$ is real-valued, so $f: \mathcal{X} \rightarrow \mathcal{R}$. $f$ is parameterized by a mean function $\mu$ and a covariance kernel function $k(\cdot,\cdot)$:
\begin{equation}
f \sim \mathcal{GP}(\mu,k(\cdot,\cdot))
\end{equation}
meaning that:
\begin{align}
E[f(x)] & = \mu(x) \\
\mbox{Cov}(f(x),f(x')) & = k(x,x')
\end{align}
The defining feature of a Gaussian process is that at any finite set of indices $x_1, \ldots, x_n \in \mathcal{X}$, the distribution of the vector $[f(x_1), \ldots, f(x_n)]^{\top}$ is a multivariate Gaussian:
\begin{equation}
	\label{eq:mvn}
[f(x_1), \ldots, f(x_n)]^{\top} \sim \mathcal{MVN}([\mu(x_1), \ldots, \mu(x_n)]^{\top}, K)
\end{equation}
where the covariance matrix $K_{ij} = k(x_i,x_j)$.

Gaussian processes are often described as a nonparametric model, since the
number of parameters grows with the sample size, with attractive posterior
consistency results in a simple regression setting \cite{choi2007posterior}, 
and as part of a Cox survival model \cite{FerTeh2016a}. 

\subsection{Scalable kernel methods} \label{section:scalable-kernel}
The matrix algebra operations required for calculations involving
multivariate Gaussians are not scalable to large datasets. Practitioners face the same
issue when applying nonlinear kernel methods \cite{scholkopf2002learning}, as
they rely on the calculation and manipulation of an $n \times n$
Gram matrix, which corresponds exactly to the covariance matrix parameterized
by the covariance kernel in Gaussian processes. While a variety of approaches have
been proposed to alleviate this computational difficulty (for a comprehensive comparison in the spatial setting, see
\citet{heaton2017case}), we consider random Fourier features \cite{rahimi2007random,rahimi2008weighted}
due to the simplicity with which it can be embedded within a larger supervised learning framework.

Random Fourier Features are a randomized approximation yielding a
finite-dimensional feature mapping (and corresponding finite-dimensional
Reproducing Kernel Hilbert Space) which approximates the original kernel.
Proposed in 2007 \cite{rahimi2007random} and marketed as only requiring 3 lines
of MATLAB code to apply, the authors won a ``Test of Time Award'' at NIPS in 2017 for
their widespread applicability and elegance.

Recall Bochner's theorem (for a precise statement in multiple dimensions
see \cite[p. 24]{stein1999interpolation}),
which establishes a one-to-one correspondence between a stationary kernel 
$k(x_i,x_j)=\kappa(x_i-x_j)$  and a positive finite measure $\Lambda$. In particular, 
$\kappa$ is the Fourier transform of $\Lambda$:
\begin{equation}
	\kappa(x-y) = \int_\mathcal{R} e^{i \omega^{\top} (x-y)} d \Lambda(\omega)
\end{equation}
Recognizing that this infinite dimensional integral is an expectation over the measure $\Lambda$ suggests
that it can be approximated using Monte Carlo sampling. After normalizing appropriately, we treat
$\Lambda$ as a pdf and consider iid samples:
\begin{equation}
	\omega_1, \ldots, \omega_d \sim \Lambda(\omega)
\end{equation}
Then we have the following approximation:
\begin{align}
	\kappa(x-y) & \approx \frac{1}{d} \sum_{j=1}^d e^{i \omega_j^{\top} (x-y)} \\
	& = \frac{1}{d} \sum_{j=1}^d \cos(\omega_j^{\top} (x-y)) + i \sin(\omega_j^{\top}(x-y)) \\\label{eq:scalable}
	& = \frac{1}{d} \sum_{j=1}^d \cos(\omega_j^{\top} x)\cos(\omega_j^{\top} y) + \sin(\omega_j^{\top} x)\sin(\omega_j^{\top} y) \\\label{eq:simplify}
	& = \langle \Phi(x), \Phi(y) \rangle
\end{align}
Since we know that the kernel is real-valued, in line \eqref{eq:scalable} we 
ignore the imaginary component and expand using the trigometric identity for $\cos(\alpha + \beta)$.
We can now define an explicit feature mapping $\Phi: \mathcal{X} \rightarrow \mathcal{R}^{2d}$ consisting
of the following pairs of elements concatenated together:
\begin{equation}
\Phi_j(x) = \frac{1}{\sqrt{d}} \left(\begin{array}{c} \cos(\omega_j^{\top}x) \\
\sin(\omega_j^{\top}x)
\end{array}
\right)
	\label{eq:features}
\end{equation}

\begin{figure}[h!]
	\centering
	\includegraphics[width=.4\paperwidth]{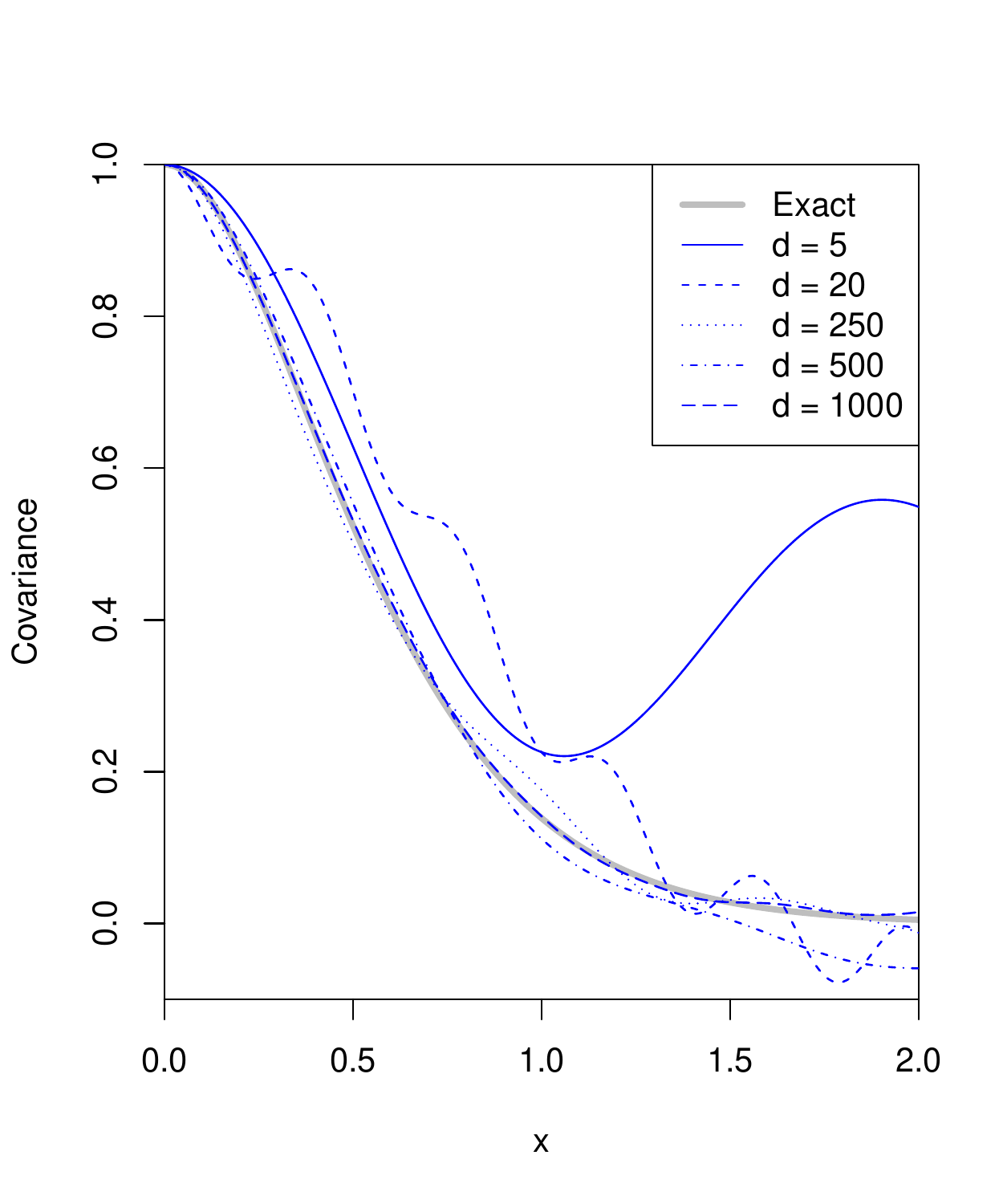}
	\caption{We approximate a Mat\'ern 5-2 with varying number of random Fourier features. The approximation becomes 
	more and more exact as the number of features increases.}
	\label{fig:rff-compare}
\end{figure}
The elegance of random Fourier features is that in a supervised learning
setting with $n$ observations and $p$ covariates
with design matrix $X \in \mathcal{R}^{n \times p}$, we can immediately consider 
introducing kernel-based nonlinearities, without changing our learning approach,
simply by transforming our design matrix into $[\cos(X \Omega^{\top}) \quad \sin(X \Omega^{\top})]$ 
for a set of random frequencies $\Omega \in \mathcal{R}^{d \times p}$ 
(here, $\cos(A)$ is the element-wise computation of cosine on $A$). 
In homage to the three lines of MATLAB code, three lines of R code are shown below to transform a design matrix {\tt X}
into a new design matrix {\tt Phi} assuming a squared exponential kernel with lengthscale 1, 
$k(x,y) = \exp(-.5\|x-y\|^2)$. In our space-time setting, {\tt X} has 3 columns giving the x, y, and t coordinates of
the observations, but it could also include covariates if available.
{\tt
\begin{verbatim}
Omega = matrix(rnorm(d*ncol(X)), d)
Proj = X %*% t(Omega)
Phi = cbind(cos(Proj), sin(Proj)) / sqrt(d)
\end{verbatim}
}
Changing the covariance kernel's lengthscale corresponds to changing the variance
of the normal distribution. Using a Mat\'ern kernel instead of a squared exponential requires 
sampling from a Student-t distribution instead of a normal distribution.
At this point, any (suitably regularized) linear learning method can be applied to {\tt Phi}:
ridge regression with {\tt Phi} is an approximation to kernel ridge regression with {\tt X}; 
Bayesian linear regression with {\tt Phi} is an approximation to Gaussian process regression with {\tt X}. 

As shown in Figure \ref{fig:rff-compare}, a larger number of random features $d$ increases the accuracy
of the approximation.

\subsection{Hawkes features vs.~KDE features}
\label{appendix:hawkes-features}

Our approach is similar to the ``Hawkes features'' used by \citet{mohler2018rotational}. 
However, as detailed below, both this previous work and ours are actually akin to the little studied
nonlinear Hawkes process, rather than the more standard linear Hawkes process.
The conditional intensity function used in a spatiotemporal linear Hawkes process takes the following form:
\begin{equation}
\lambda(x,y,t) = \lambda_0(x,y,t) + \sum_{\{i | t_i < t\}} k_t(t_i,t) \cdot k_s((x_i,y_i),(x,y))
	\label{eq:hawkes}
\end{equation}
where the first term is an underlying (endogeneous) intensity and the second term is the self-excitatory
component.

We argue that in the supervised learning framework of training a model to
predict the future given the present, the main distinction between an unweighted KDE
and the Hawkes process disappears. The reason is that in the supervised framework, the
lagged KDE features only have access to past events, so the intensity cannot
rise before an event occurs, i.e.~the directionality of time is enforced. 

As defined in Eq.~\ref{eq:kde}, $\mbox{KDE}_{\lambda,1}(x,y,t)$ is the lag-1 spatial kernel density estimator at location $(x,y)$ using data with time labels $\in [t - D, t]$, i.e.:
\begin{equation}
\mbox{KDE}_{\lambda,1}(x,y,t) = \sum_{\{i | t_i \in [t-D,t]\}} k_s((x_i,y_i),(x,y))
	\label{eq:kde1}
\end{equation}

If we consider Eq.~\eqref{eq:hawkes} and the special case in which 
\begin{equation}
	k_t(t_i,t) = 
\begin{cases}
	1 ~\mbox{if}~  t_i \in [t-D,t]\\
	0 ~ \mbox{otherwise}
\end{cases}
	\label{eq:boxcar}
\end{equation}
the KDE feature is equivalent to the self-excitatory term in the Hawkes process conditional
likelihood. 
This result makes sense in a supervised learning framework, in which the KDE values are computed ignoring future data (from the point of view
of the features, the future has not occurred yet).

Note, however, beyond the particular kernel choice and lack of a time
component, there is another major distinction between prior and present work:
the log link function in our GLM framework implies that instead of these features contributing
additively to the intensity, the exponential of their sum contributes to the intensity, or equivalently
the product of exponentials.
This is analogous to the rather exotic ``nonlinear'' Hawkes process
\cite{gerhard2017stability,carstensen2010hawkes,zhu2013central}, where the
effect of past events on the present intensity is multiplicative, rather than
additive, and for which stability results are not well-established.  This
distinction applies to \citet{mohler2018rotational} as well, due to the
inclusion of the logistic link function.

The kernel in Eq.~\eqref{eq:boxcar} is rather simplistic, 
suggesting an obvious extension to our method of including a more interesting
temporal kernel in the KDE features of Eq.~\eqref{eq:kde1} and possibly also for other lags as in Eq.~\eqref{eq:kde}.
It would be sensible as well to consider the more standard linear Hawkes formulation, which is known to be stable,
instead of the nonlinear version considered here.

\newpage

\section{Supplementary Results}
\begin{figure}[ht!]
\centerline{
	\includegraphics[width=1.1\textwidth]{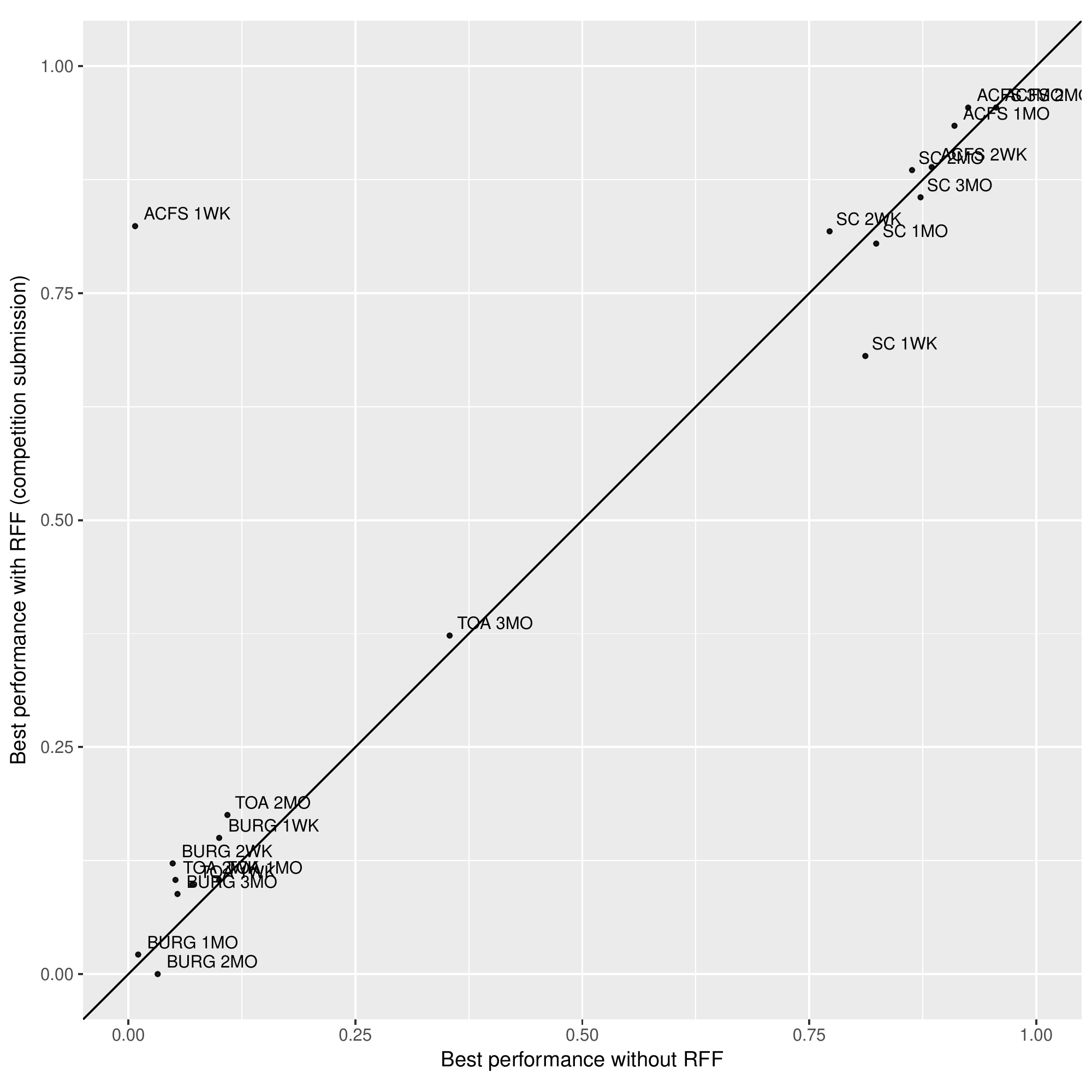}}
    \caption{Model Omitting Random Fourier Features (x-axis) compared to Full Model (y-axis). The Full Model out-performed the Model Omitting Random Fourier Feature in 15 out of 20 forecast problems. The average out-of-sample performance improvement of the Full Model over the Non-Rff Model was 0.05 on the PEI scoring metric. BURG = burglary, SC = street crime, TOA = theft of auto, ACFS = all calls for service. }
    \label{fig:sub2}
\end{figure}
\begin{figure}[hb!]
	\centerline{
    \includegraphics[width=1.25\textwidth]{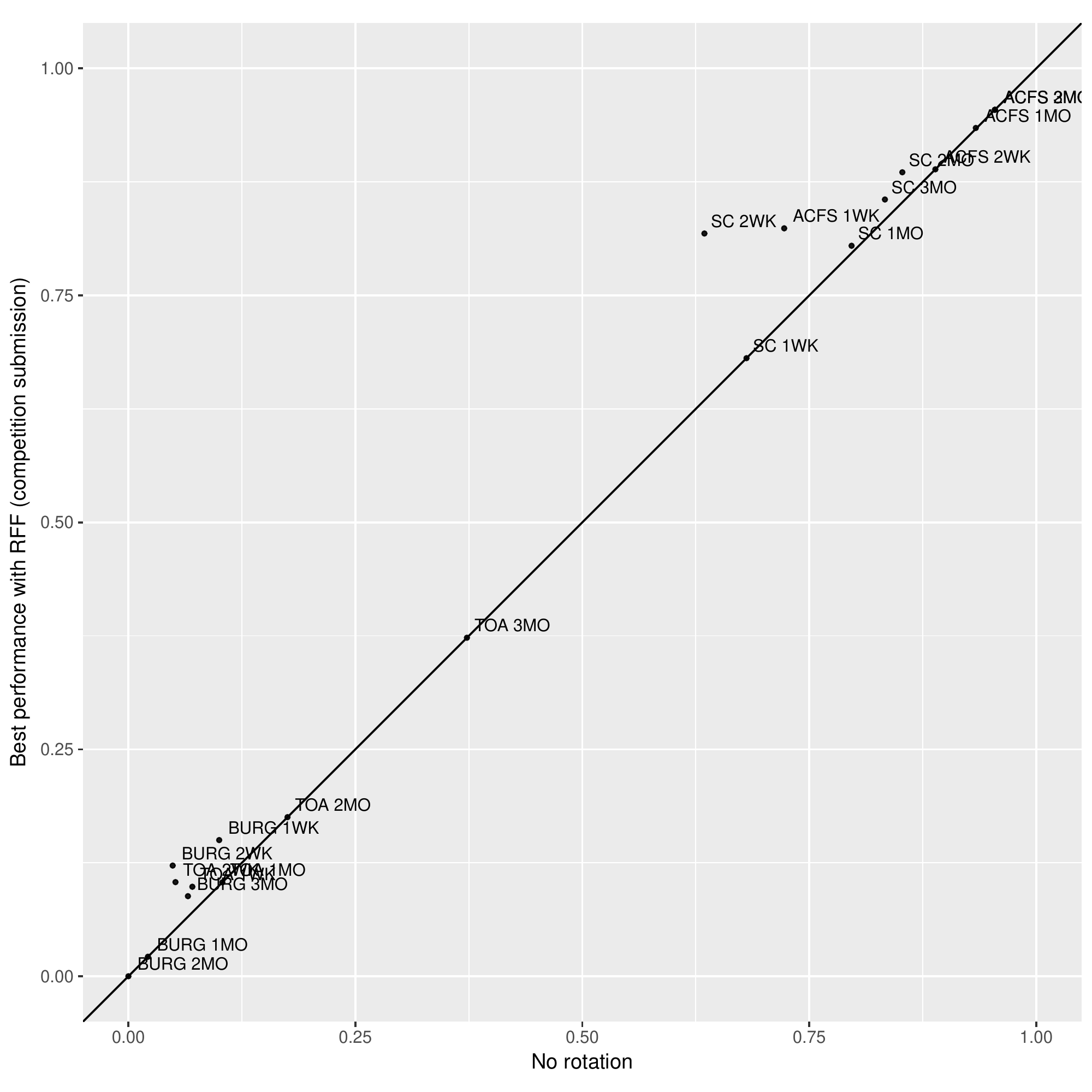}}
	\caption{Model Omitting Grid Rotation (x-axis) compared to Full Model (y-axis). The Full Model out-performed the Model Omitting Grid Rotation in 10 out of 20 forecast problems. The average out-of-sample performance improvement of the Full Model over the Non-Rotated Model was 0.029 on the PEI scoring metric. BURG = burglary, SC = street crime, TOA = theft of auto, ACFS = all calls for service. }
    \label{fig:sub3}
\end{figure}

\clearpage
\begin{figure}[ht!]
\centerline{
    \includegraphics[width=1.25\textwidth]{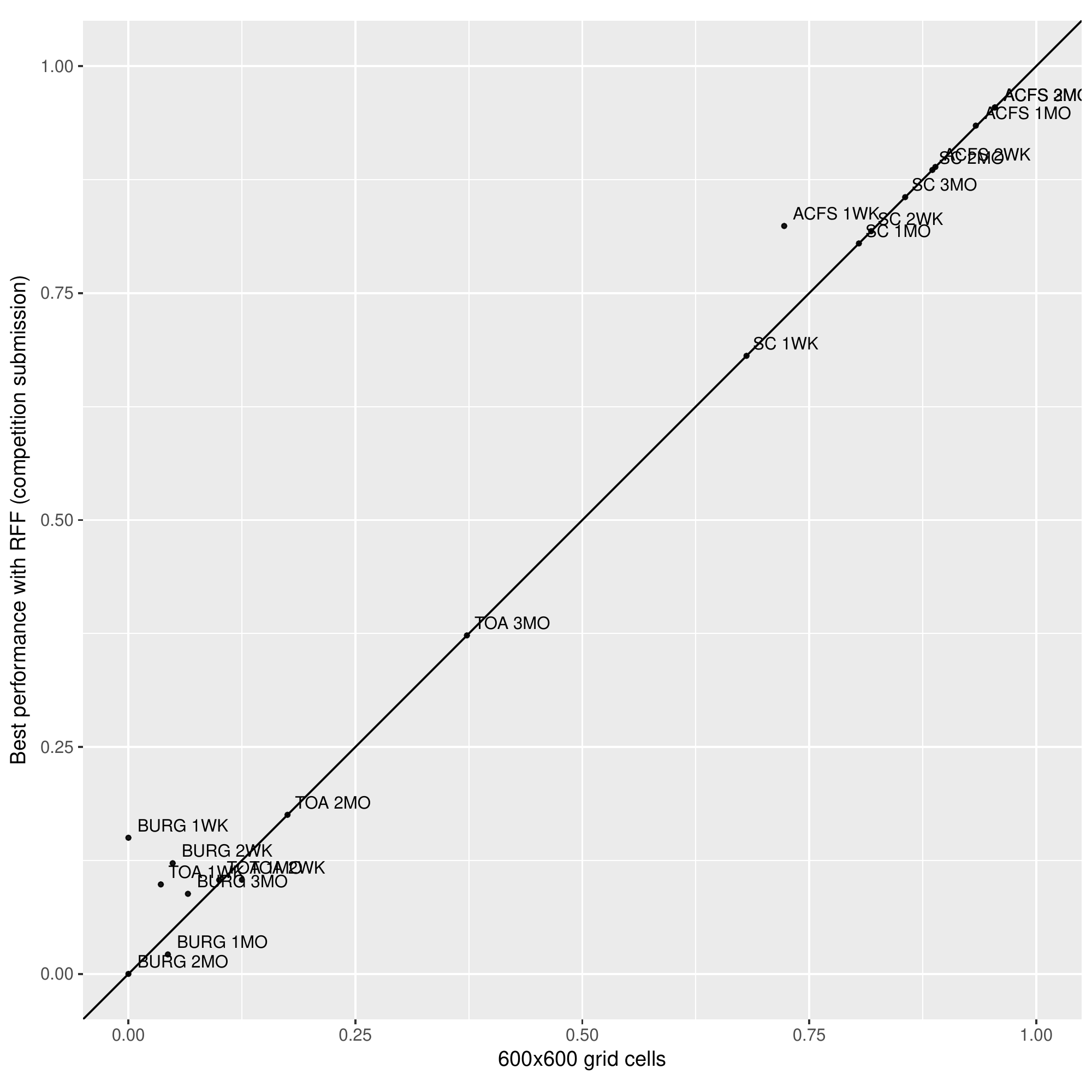}}
	\caption{Model with Fixed Grid Cells (x-ais) compared to Full Model (y-axis). The Full Model out-performed the Model with Fixed 600x600 ft cells in 6 out of 20 forecast problems. The average out-of-sample performance improvement of the Full Model voer the Fixed Cell Model was 0.019 on the PEI scoring metric. BURG = burglary, SC = street crime, TOA = theft of auto, ACFS = all calls for service. }
	\label{fig:sub4}
\label{fig:additional}
\end{figure}

\clearpage
	\section{Hyperparameter choice}
	\label{appendix:hypers}

Portland, like many cities, has a mix of north/south and east/west aligned streets. However, it also has a non-trivial number of obliquely-oriented streets and parcels. This is especially the case in downtown Portland, east of the Williamette River and south of West Burnside Street. 
Given the concentration of calls for service in this area, it seemed likely that a non-standard alignment could be beneficial, especially for all calls for service. For this reason, grid angle of rotation was included as another parameter to be learned by  the model, an idea first proposed by \cite{johnson_predictive_2009}.
\begin{figure}[ht!]
	\centering
	\includegraphics[width=.5\paperwidth]{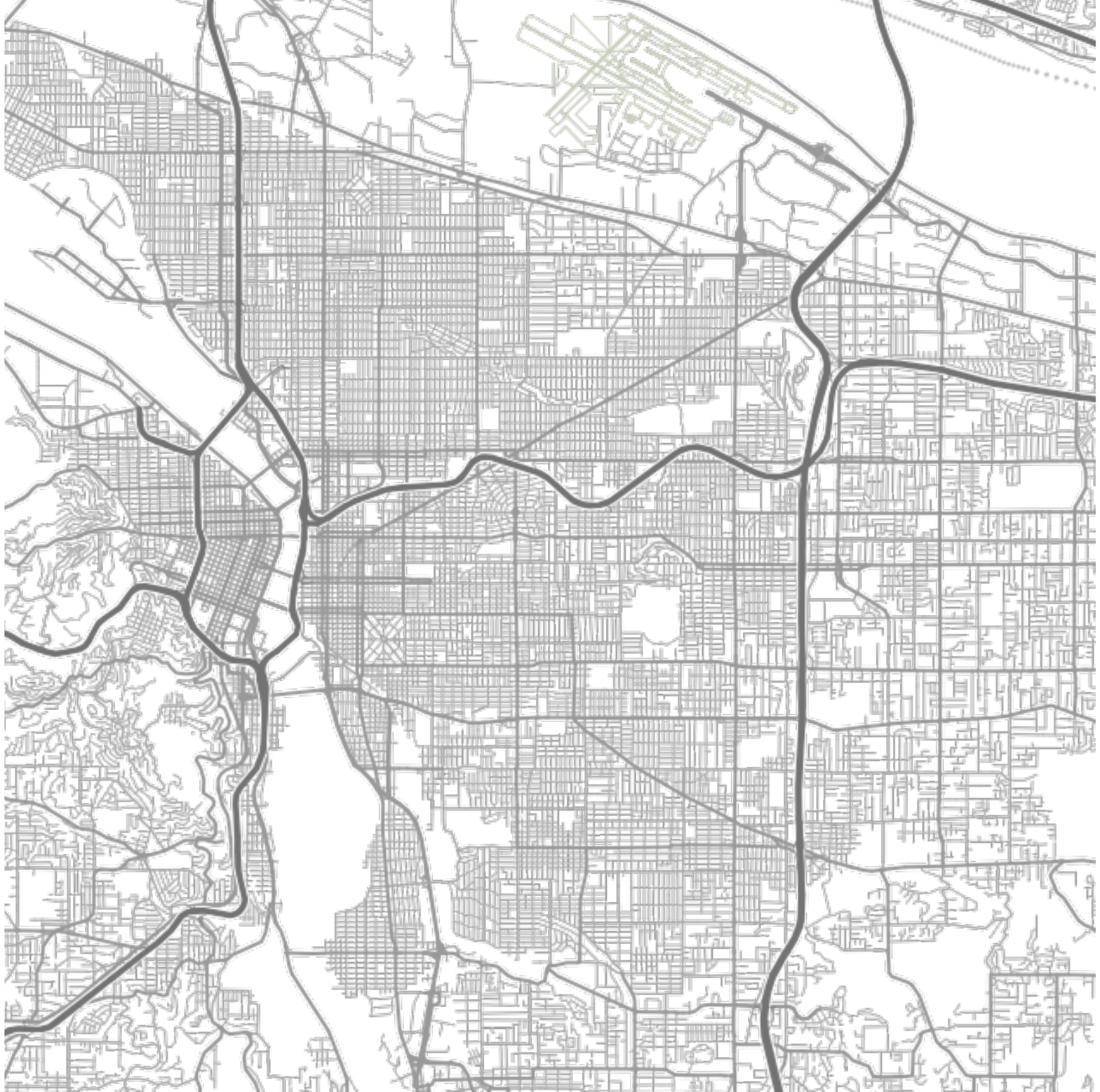}
	\caption{Central Portland}
	\label{fig:Portland}
\end{figure}
Non-uniform land use coupled in Portland with the common practice of geocoding crime data to the street grid suggested that a fixed N/S oriented square tessellation could be sub-optimal \cite{Rosser_predictive_2016}.\footnote{Data provided by NIJ included calls geocoded to the building footprints and then offset several feet onto the street grid directly in front of the relevant address.} However, contest rules required that shapes be polygons that could be tesselated without rotation. While squares were the simplest choice, the nature of the calls for service data geocoded to the street grid suggested that rectangles could be preferable\footnote{We also considered the other regular tesselations of the plane -- namely, equilateral triangles and regular hexagons. Both suffered from computational problems, as no extant open libraries offer scalably-fast kernel density estimation over non-rectangular polygons. Nevertheless, some firms (notably Uber) use hexagons at scale for spatiotemporal forecasting.}. For this reason, we chose to leave the cell shape as a parameter to be optimized with each crime type and forecasting period potentially receiving its own optimized solution. Similarly, contest rules permitted a variety of different cell sizes and consistent with recent working demonstrating the sensitivity of forecast accuracy to cell size \cite{timothy_hart_kernel_2014}, we left this as a parameter to be optimized. Subject to processing resource limitations, theoretically, any parameter could be optimized for forecasting accuracy.
	
The final hyperparameters we selected are shown in Table \ref{table:hypers}. Winning entries are highlighted in yellow.

\begin{landscape}

	\pagestyle{empty}
	\begin{table}
\centerline{
	\begin{tabular}{|p{1.4cm}|p{1.0cm}|p{1.2cm}|p{1.0cm}|p{1.2cm}|p{1.1cm}|p{1.2cm}|p{1.0cm}|p{1.0cm}|p{1.0cm}|p{1.0cm}|p{1.0cm}|p{1.0cm}|p{1.5cm}|}
  \hline
		Horizontal grid size (ft) & Vertical grid size (ft) & Coverage area & Spatial lengthscale (ft) & Temporal lengthscale (days) & Rotation angle (radians) & Number of random features $d$ & l1 regularization & l2 regularization & KDE bandwidth (ft) & Number of KDE lags & KDE window (days) & Crime Type & Forecasting Period \\
  \hline
	\rowcolor{yellow} 478 & 710 & 10\% & 570 & 67.10 & 0.85 & 362 & 0 & 5e-5 & 274.70 & 9 & 39.62 & ACFS & 1m \\\hline
	\rowcolor{yellow}  618 & 473 & 16\% & 457.5 & 42.93 & 0.25 & 360 & 0 & 0 & 391.97 & 9 & 68.73 & ACFS & 1w \\\hline
  600 & 600 & 0\% & 250 & 60 & 0 & 250 & 0 & 1e-5 & 500 & 8 & 45 & ACFS & 2m \\\hline
  600 & 600 & 0\% & 250 & 28 & 0 & 250 & 0 & 5e-4 & 500 & 12 & 45 & ACFS & 2w \\\hline
	\rowcolor{yellow}   600 & 600 & 5\% & 500 & 90 & 0 & 20 & 1e-5 & 1e-4 & 500 & 6 & 90 & ACFS & 3m \\\hline
  250 & 250 & 95\% & 125 & 60 & 0 & 5 & 0 & 5e-5 & 250 & 12 & 15 & burglary & 1m \\\hline
	\rowcolor{yellow}   250 & 250 & 95\% & 750 & 7 & 0 & 20 & 0 & 0 & 250 & 6 & 10 & burglary & 1w \\\hline
  431 & 598 & 10\% & 1250 & 18.95 & 0 & 10 & 0 & 1e-4 & 342.42 & 10 & 3.50 & burglary & 2m \\\hline
	\rowcolor{yellow}   250 & 250 & 100\% & 125 & 70 & 0 & 5 & 0 & 1e-5 & 250 & 6 & 21 & burglary & 2w \\\hline
  689 & 484 & 15\% & 847.5 & 105.96 & 0.37 & 36 & 0 & 0 & 597.12 & 4 & 14.77 & burglary & 3m \\\hline
  600 & 600 & 15\% & 250 & 15 & 0.98 & 20 & 0 & 0 & 500 & 3 & 15 & street & 1m \\\hline
  600 & 600 & 10\% & 125 & 3.50 & 0 & 250 & 0 & 0 & 500 & 6 & 7 & street & 1w \\\hline
  600 & 600 & 10\% & 375 & 120 & 0.98 & 20 & 0 & 0 & 500 & 1 & 60 & street & 2m \\\hline
	\rowcolor{yellow}   600 & 600 & 5\% & 125 & 7 & 0.98 & 250 & 0 & 0 & 500 & 3 & 14 & street & 2w \\\hline
  600 & 600 & 0\% & 500 & 90 & 1.18 & 20 & 0 & 5e-4 & 500 & 3 & 45 & street & 3m \\\hline
  800 & 450 & 0\% & 125 & 150 & 0 & 5 & 0 & 5-e4 & 500 & 3 & 15 & auto & 1m \\\hline
	\rowcolor{yellow}   250 & 250 & 95\% & 500 & 49 & 0 & 5 & 0 & 0 & 250 & 6 & 10 & auto & 1w \\\hline
  600 & 600 & 80\% & 125 & 60 & 0 & 20 & 0 & 0 & 500 & 1 & 30 & auto & 2m \\\hline
  250 & 250 & 100\% & 750 & 14 & 0 & 5 & 0 & 0 & 250 & 3 & 21 & auto & 2w \\\hline
  600 & 600 & 0\% & 125 & 180 & 0 & 20 & 0 & 1e-5 & 500 & 3 & 45 & auto & 3m \\\hline
\end{tabular}}
		\caption{Final hyperparameters submitted to the competition. Yellow are winning rows for the PEI metric. 
		We also won for the PAI metric for burglary 1 week and 2 weeks.
		Coverage area is a percentage of the range of areas allowed in the competition:
		0\% is the minimum (0.25 sq miles) and 100\% is the maximum (0.75 sq miles).
		ACFS = all calls for service, street = street crime, auto = theft of auto.
		}
		\label{table:hypers}
	\end{table}
\end{landscape}

\clearpage

\end{document}